\documentclass[11pt]{article}

\usepackage{amsmath}      
\usepackage{amssymb}      
\usepackage{graphicx}     
\usepackage{booktabs}     
\usepackage{multirow}     
\usepackage{hyperref}     
\usepackage{textcomp}
\usepackage[margin=1in]{geometry}  

\usepackage{tikz}
\usetikzlibrary{positioning, shapes.geometric, arrows.meta}

\tikzset{
  block/.style = {rectangle, draw, fill=blue!20, text width=7em, align=center, rounded corners, minimum height=3em},
  smallblock/.style = {rectangle, draw, fill=blue!10, text width=5em, align=center, rounded corners, minimum height=2.5em},
  arrow/.style = {thick, ->, >=Stealth}
}

\title{Scalable Unit Harmonization in Medical Informatics via \\Bayesian-Optimized Retrieval and \\Transformer-Based Re-ranking}

\author{
	\href{https://orcid.org/0000-0002-8142-7983}%
	{\includegraphics[scale=0.06]{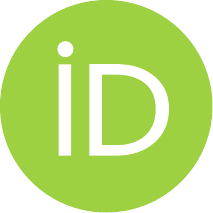}\hspace{1mm}Jordi de la Torre} \\
	Data Science \& Artificial Intelligence, Biopharma R\&D \\
	AstraZeneca \\
	Avinguda Diagonal, 615 \\
	08028 Barcelona, Spain \\
	Email: \texttt{jordi.delatorre@astrazeneca.com}
}

\date{May 1, 2025}

\begin{document}

\maketitle

\begin{abstract}
\textbf{Objective:} To develop and evaluate a scalable methodology for harmonizing inconsistent units in large-scale clinical datasets, addressing a key barrier to data interoperability.

\textbf{Materials and Methods:} We designed a novel unit harmonization system combining BM25, sentence embeddings, Bayesian optimization, and a bidirectional transformer based binary classifier for retrieving and matching laboratory test entries. The system was evaluated using the Optum Clinformatics Datamart dataset (7.5 billion entries). We implemented a multi-stage pipeline: filtering, identification, harmonization proposal generation, automated re-ranking, and manual validation. Performance was assessed using Mean Reciprocal Rank (MRR) and other standard information retrieval metrics.

\textbf{Results:} Our hybrid retrieval approach combining BM25 and sentence embeddings (MRR: 0.8833) significantly outperformed both lexical-only (MRR: 0.7985) and embedding-only (MRR: 0.5277) approaches. The transformer-based reranker further improved performance (absolute MRR improvement: 0.10), bringing the final system MRR to 0.9833. The system achieved 83.39\% precision at rank 1 and 94.66\% recall at rank 5.

\textbf{Discussion:} The hybrid architecture effectively leverages the complementary strengths of lexical and semantic approaches. The reranker addresses cases where initial retrieval components make errors due to complex semantic relationships in medical terminology.

\textbf{Conclusion:} Our framework provides an efficient, scalable solution for unit harmonization in clinical datasets, reducing manual effort while improving accuracy. Once harmonized, data can be reused seamlessly in different analyses, ensuring consistency across healthcare systems and enabling more reliable multi-institutional studies and meta-analyses.
\end{abstract}

\section{Background and Significance}

Modern clinical research increasingly relies on integrating heterogeneous data sources, yet inconsistent units of measurement remain a persistent challenge, particularly in laboratory analyses where identical quantities may be reported using diverse conventions. Harmonizing these units is essential for large-scale studies, enabling interoperability across electronic health record systems and supporting reliable, reproducible analyses. Traditional approaches often rely on manually curated rules and mapping tables developed by domain experts \cite{cholan2022encoding}. While effective in limited contexts, these methods are labor-intensive, difficult to scale, and require ongoing maintenance to accommodate new laboratory codes and units, introducing bottlenecks and delaying research workflows.

Recent advances in machine learning and information retrieval provide promising avenues for automated unit harmonization. Techniques such as sentence embeddings, bidirectional transformers, and optimized retrieval models have demonstrated improved scalability, adaptability, and contextual understanding \cite{askari2023injecting, chen2024bge, yang2025robust, 10774454}. In this study, we present a novel automated unit harmonization system that integrates Bayesian-optimized BM25, sentence embeddings, and transformer-based re-ranking. This framework addresses major limitations of existing approaches, including poor scalability, limited adaptability to diverse naming conventions, and overreliance on manual validation, while incorporating dynamic feedback mechanisms.

Beyond enabling seamless data integration, harmonizing units has a direct impact on downstream clinical and research analytics. By consolidating columns representing the same measurement, redundancy is reduced, the number of observations per feature increases, and correlations between duplicate columns are eliminated. These improvements enhance statistical power, increase the reliability of feature importance scores, facilitate detection of associations, and reduce noise from inconsistent measurements. Consequently, unit harmonization strengthens the interpretability and robustness of analytical models, supporting more reliable, timely, and actionable scientific insights that can ultimately inform clinical decision-making.

Current approaches to unit harmonization span a spectrum from basic open-source tools to sophisticated commercial platforms. Open-source frameworks such as ehrapy \cite{heumos2024open}, psHarmonize \cite{stephen2024psharmonize}, and MIMIC-Extract \cite{wang2020mimic} provide foundational harmonization capabilities but typically require substantial manual intervention and offer limited scalability for large datasets. Commercial platforms deliver advanced automation and can handle large-scale data processing, but their proprietary nature limits accessibility and customization for research applications.

Our proposed framework addresses this gap by combining Bayesian-optimized BM25, sentence embeddings, and transformer-based re-ranking to achieve advanced automation and scalability comparable to commercial systems while maintaining the flexibility and accessibility of open-source solutions. This hybrid approach enables the processing of billions of records with minimal manual intervention, providing both robustness and practical applicability for large-scale clinical research. Importantly, the system is designed with a modular architecture that makes it easily extendable to any alternative dataset, supporting diverse clinical coding standards including Loinc \cite{mcdonald2003loinc}, ATC \cite{nahler2009anatomical}, MedDRA \cite{brown1999medical}, RxNorm \cite{liu2005rxnorm} and SNOMED-CT \cite{donnelly2006snomed} beyond the unit harmonization focus of this study.

\section{Materials and Methods}

\subsection{Datasets}

\subsubsection{Primary Dataset}

The Optum Clinformatics Data Mart (CDM) \cite{optumclinformatics2024} constitutes our primary evaluation dataset. This comprehensive, de-identified database integrates administrative health claims data from over 84 million individuals across all 50 U.S. states. The resource contains more than 7.5 billion medical and pharmacy claims, documenting healthcare utilization and associated costs. Data elements include member demographics, detailed medical and pharmacy claims, laboratory results, inpatient confinement records, and provider information.

The large scale and heterogeneity of the dataset provide a rigorous testbed for evaluating the effectiveness and scalability of our unit harmonization system. The vast diversity of lab tests, samples, and reported units reflects real-world variability and complexity in clinical data sources, posing significant challenges for accurate and automated harmonization. Furthermore, the inclusion of data from diverse healthcare settings across all U.S. states ensures that the system is evaluated against a wide range of measurement practices and data quality variations.

Given that Optum CDM is extensively used for pharmacoepidemiology, outcomes research, and healthcare analytics, successful harmonization of units within this dataset directly contributes to improving the reliability and consistency of downstream clinical analyses and decision-making. Therefore, demonstrating our system’s performance on this dataset not only validates its technical robustness but also its practical utility in large-scale medical informatics applications.

In this study, we used the 2024 Q1, Q2, and Q3 version of the Optum Clinformatics Data Mart. To support the harmonization process, key fields were extracted from the dataset, including LOINC codes, units of measurement, frequency of occurrence, and descriptive statistics such as minimum, maximum, mean, and standard deviation. This process resulted in approximately 30,000 unique triads (test, sample, unit) requiring standardization. Of these, 17,244 entries were identified in our internal database as candidates for matching.

\subsubsection{Reference Dataset}

We utilized an internal Labcodes Standard Database for unit harmonization, providing a comprehensive mapping of laboratory tests with standardized information. The database facilitates consistent data representation across diverse sources by linking original units to preferred units and providing necessary conversion factors.

The harmonization process is mapped against multiple fields within this database, including Test Name, Test Label, Synonym, Sample Name, Labcode, Preferred Unit, Actual Unit, Conversion Factor, and various CDISC standardization fields. Although all fields contribute to a comprehensive test definition, harmonization can often be achieved with a minimum of Test Name, Sample Name, and Unit. Inclusion of additional fields improves the accuracy and specificity of the match.

\subsection{System Architecture}

Our harmonization framework implements a multi-stage pipeline comprising data processing, predictive modeling, and validation components. A key design principle is the modular decomposition of this framework into independent blocks, allowing for focused development, rigorous testing, and iterative improvement of individual components.

\subsubsection{Overall Pipeline Structure}

\begin{figure}[ht]
\centering
\includegraphics[width=0.9\textwidth]{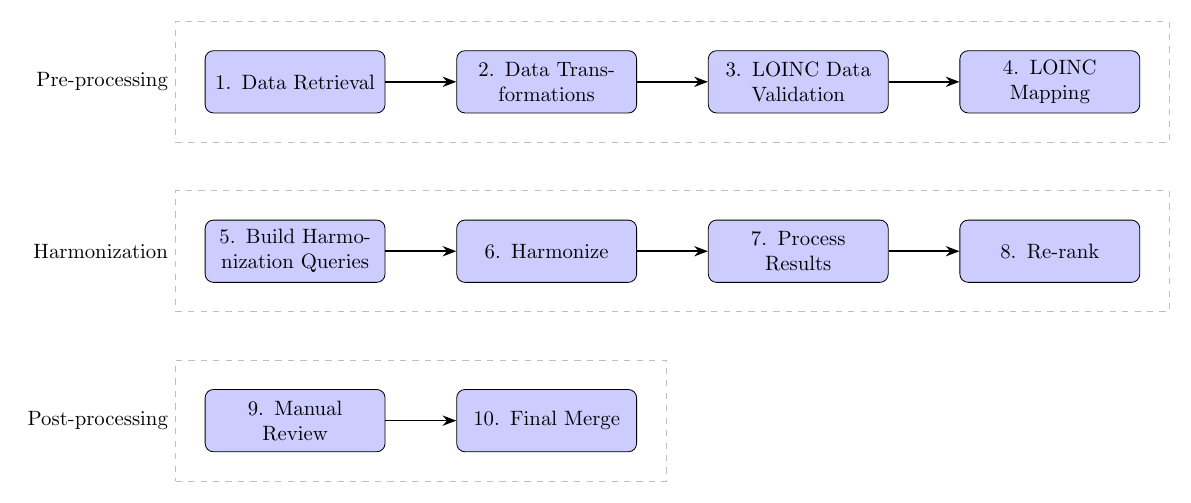}
\caption{Overview of the harmonization pipeline showing the sequential flow of data from raw database entries to validated harmonized units}
\label{fig:pipeline}
\end{figure}

The system is organized as an end-to-end processing pipeline (Fig.~\ref{fig:pipeline}) comprising three main stages: preprocessing, harmonization, and post-processing. The workflow begins with preprocessing, which retrieves raw data from the source system, Optum Clinformatics Data Mart. This stage focuses on relevant elements such as LOINC codes—standardized identifiers for laboratory tests—and reported units. The data then undergoes cleaning and normalization to ensure consistency. A validation step follows, confirming the integrity of LOINC codes and associated metadata. Finally, the dataset is enriched by mapping LOINC codes to standardized labels and attributes, adding valuable information to support the subsequent harmonization phase.

During harmonization, the system constructs tailored queries that generate ranked proposals for matching equivalents. These queries leverage the hybrid retrieval engine presented in this paper, enabling efficient searches for equivalent data elements. The hybrid retrieval engine returns candidate matches—potential harmonization targets retrieved from the reference database. These represent the initial pool of harmonization proposals returned by the system based on lexical (BM25) and semantic similarity scores, before refinement through the transformer-based reranking process. The retrieved results are then processed and structured for evaluation, followed by a reranking step that prioritizes candidates based on relevance scores to enhance overall accuracy.

The pipeline concludes with post-processing, where domain experts review the harmonization outcomes through a custom interface to ensure clinical safety and correctness. Finally, validated results are integrated into the target repository, completing the harmonization process.

As illustrated in Fig.~\ref{fig:pipeline}, this comprehensive workflow balances automated methods with expert oversight to deliver reliable and high-quality data harmonization.

\subsubsection{Predictive Model Architecture}
\label{sec:pred-model-arch}

\begin{figure}[ht]
\centering
\includegraphics[width=0.9\textwidth]{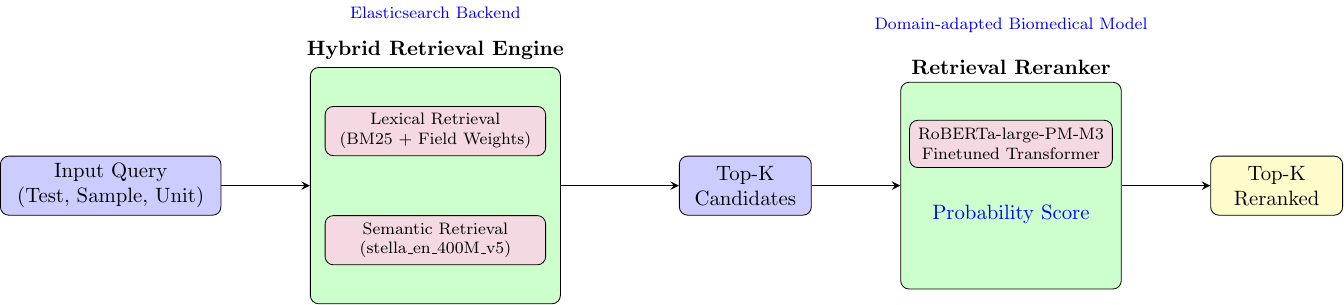}
\caption{Components of the predictive model architecture showing the interaction between retrieval mechanisms and contextual reranking}
\label{fig:architecture}
\end{figure}

The core intelligence of the system is embodied in the predictive model architecture (Fig.~\ref{fig:architecture}), which consists of a \textbf{Hybrid Retrieval Engine} and a \textbf{Retrieval Reranker}.

The Hybrid Retrieval Engine combines two complementary retrieval methods: the Lexical Retrieval Module, which uses the BM25 algorithm for precise keyword matching and excels at exact and fuzzy matches of laboratory terminology; and the Semantic Retrieval Module, which employs sentence embeddings to capture contextual relationships, enabling detection of semantically equivalent terms even when lexical overlap is limited.

A key design feature within this engine is the \textit{Bayesian Score Optimizer}, which uses Gaussian Process-based optimization to determine the optimal weighting between lexical and semantic scores. This optimization is performed offline before inference, ensuring that precomputed weights guide the weighted combination of retrieval scores during query execution.

The Retrieval Reranker serves as the final decision layer. This bidirectional transformer model reevaluates and reorders candidate matches based on deeper contextual understanding, addressing subtle semantic distinctions critical in complex medical terminology.

The retrieval engine relies on an Elasticsearch backend, supporting scalable and flexible querying via JSON schema definitions that govern indexing and search of laboratory metadata. Candidate ranking scores are calculated as a weighted linear combination of BM25 lexical scores and semantic similarity scores measured by cosine distance between embeddings, with weights established by the Bayesian optimization design embedded within the Hybrid Retrieval Engine.

This modular architecture achieves both high accuracy and scalability. Broad candidate coverage is provided by the hybrid retrieval engine, while the reranker ensures precision through deeper semantic analysis. Each component can be independently optimized and updated, maintaining backward compatibility and allowing incorporation of future advances.

\subsubsection{Baseline Retrieval Using BM25}

Our retrieval system incorporates as a base the BM25 algorithm \cite{robertson2009probabilistic}, a state-of-the-art probabilistic relevance framework. BM25 offers a refined modeling of term frequency, along with document length normalization to mitigate frequency bias. The BM25 score for a document \(d\) with respect to a query \(q\) is computed as:

\begin{equation*}
\text{score}(d, q) = \sum_{t \in q} \text{IDF}(t) \cdot \frac{\text{TF}(t, d) \cdot (k_1 + 1)}{\text{TF}(t, d) + k_1 \cdot \left(1 - b + b \cdot \frac{|d|}{\text{avgdl}}\right)}
\end{equation*}

where:

\begin{itemize}
    \item $\text{TF}(t,d)$ is the term frequency of term $t$ in document $d$
    \item $\text{IDF}(t)$ is the inverse document frequency of term $t$
    \item $|d|$ is the length of document $d$ in words
    \item $\text{avgdl}$ is the average document length in the collection
    \item $k_1$ (typically 1.2-2.0) and $b$ (typically 0.75) are free parameters that control term frequency scaling and document length normalization respectively
\end{itemize}

Our system builds on the strengths of the BM25 ranking function to support both exact and fuzzy matching, optimizing retrieval performance across varying levels of lexical precision. In scenarios requiring exact matching, BM25 prioritizes documents that contain query terms with higher frequency and in close proximity. This capability is particularly beneficial in the context of laboratory code harmonization, where precise terminology—such as LOINC codes or standardized unit expressions—is critical for accurate identification and mapping.

In addition to exact matches, the system supports fuzzy matching to accommodate natural variations in terminology, including differences in spelling, formatting, or word choice. BM25 introduces a graded penalty for such partial matches, allowing relevant documents with similar but non-identical terms to be retrieved while preserving ranking quality. This is especially important in healthcare data, where naming conventions can vary across institutions or over time.

To enhance retrieval effectiveness, we incorporated curated synonym lists that expand the query space by mapping equivalent or related terms. These lists were manually developed by the primary author to address the extensive variability in unit and sample nomenclature found in clinical datasets. For laboratory units, the synonym mappings capture notation variants such as scientific notation formats (10*3/L, 10**3/L, 10\textasciicircum{}3/L), word-based expressions (THOUSAND/L, THOU/L), and symbolic alternatives (x10(3)/L, k/uL). For samples, they consolidate semantically equivalent terms such as “plasma, blood plasma, plas” and “serum, blood serum, ser,” while also resolving more complex combinations like “serum/plasma, serum or plasma, plasma/serum.” In total, the resource comprises approximately 300 unit-based and 50 sample-based synonym groups. These synonym mappings were implemented using Elasticsearch’s standard synonym filter, which integrates seamlessly with the BM25 framework by automatically expanding queries during indexing and search. Collectively, this strategy enables the retrieval engine to provide robust, flexible, and clinically meaningful harmonization of laboratory units.

\subsubsection{Contextual Retrieval via Sentence Embeddings}

Sentence embeddings \cite{reimers2019sentence,gao2021simcse,chen2020simple} serve as the semantic backbone of our retrieval system, enabling context-aware matching between queries and candidate records that go beyond lexical overlap. These embeddings are dense vector representations of text, where semantically similar sentences are mapped to nearby points in the embedding space. In the context of laboratory terminology, this is crucial for recognizing variant expressions of the same clinical concept, such as “blood urea nitrogen” and “BUN,” or for disambiguating polysemous terms based on surrounding context.

\subsubsection{Hybrid Lexical–Semantic Retrieval and Ranking}

Our hybrid search system integrates both lexical and semantic search capabilities through a multi-tiered architecture (Fig.~\ref{fig:architecture}) designed to enhance retrieval accuracy and contextual relevance. The first component of this architecture involves the generation of combined text embeddings that represent laboratory test descriptions alongside relevant metadata. These embeddings enable the system to capture nuanced semantic relationships between queries and candidate entries, going beyond surface-level term matching.

To refine retrieval precision, the system employs field-specific search boosts that reflect the relative importance of different attributes. For instance, fields such as standardized test names or unit designations may carry more weight in the matching process than auxiliary metadata. These boost factors are configurable and allow the system to dynamically adjust the influence of each field depending on the context of the search. The main sources of error in the harmonization process stem from inconsistencies in three key components: test name, sample, and unit. These elements are essential for achieving correct matches, and discrepancies in any of them can lead to harmonization errors. To address this challenge, the model can be adjusted by optimizing the relative importance (weights) of each parameter for the specific training dataset, allowing better adaptation to dataset-specific characteristics and reducing potential mismatches.

Formally, the final ranking score for a candidate document $d$ given a query $q$ is computed as a weighted combination of lexical and semantic components:

\begin{equation}
\text{Score}(d,q) = \alpha \cdot \text{BM25}(d,q) + \beta \cdot \max(0, \text{CosSim}(E_q, E_d))
\end{equation}

where:
\begin{itemize}
    \item $\alpha$ and $\beta$ are optimized weights for lexical and semantic components, respectively,
    \item $\text{BM25}(d,q)$ is the BM25 score as defined in the previous section,
    \item $\text{CosSim}(E_q, E_d)$ is the cosine similarity between query embedding $E_q$ and document embedding $E_d$,
    \item $\max(0, \cdot)$ clips negative similarities to zero, ensuring only positive semantic matches contribute.
\end{itemize}

For field-specific queries, the lexical component is further decomposed to explicitly account for the importance of harmonization-critical attributes:

\begin{equation}
\text{BM25}(d,q) = \sum_{f \in \{\text{test}, \text{sample}, \text{unit}, ...\}} w_f \cdot \text{BM25}_f(d_f, q_f)
\end{equation}

where $w_f$ represents the weight for field $f$, and $\text{BM25}_f(d_f, q_f)$ is the BM25 score computed on the field-specific content. This formulation allows the system to explicitly emphasize the test, sample, and unit fields, which are the primary sources of harmonization errors.

Structured queries are constructed to leverage both exact and fuzzy matching mechanisms, alongside semantic vector similarity. A custom-built module parses incoming queries using field-aware syntax, applies the predefined boost factors, and computes a composite score according to the equations above. By merging symbolic (BM25) and contextual (semantic embeddings) strategies in a principled way, the system achieves high recall and precision in laboratory unit harmonization tasks, even in the presence of terminological variation and incomplete data.

\subsubsection{Tuning Lexical and Semantic Weights via Bayesian Optimization}

To determine the optimal combination of retrieval methods, we implemented Bayesian optimization to fine-tune the weights assigned to various components of the search process. This approach systematically explores the parameter space, ensuring that the retrieval system operates at peak efficiency while balancing competing factors. Specifically, the optimization process focuses on adjusting field boost weights, which control the relative importance of different fields (e.g., test descriptions, specimen types, and units) in the retrieval process. By tuning these parameters, the system can adapt to the specific requirements of each search context.

Formally, the Bayesian optimization seeks to identify the optimal parameter vector
\[
\theta = [\alpha, \beta, w_{\text{test}}, w_{\text{sample}}, w_{\text{unit}}, ...]
\]
that maximizes the Mean Reciprocal Rank (MRR) on a validation set:
\begin{equation}
\theta^* = \arg\max_{\theta} \text{MRR}(\theta) = \arg\max_{\theta} \frac{1}{|Q|} \sum_{q \in Q} \frac{1}{\text{rank}_q(\theta)},
\end{equation}
where $Q$ is the set of validation queries and $\text{rank}_q(\theta)$ is the rank of the correct answer for query $q$ under parameter configuration $\theta$.

The optimization primarily targets the main sources of harmonization errors, which arise from inconsistencies in test names, sample types, and units. By systematically adjusting the weights of these critical parameters, the system minimizes mismatches and improves overall harmonization accuracy. This data-driven approach enables the retrieval model to adapt to dataset-specific characteristics, automatically learning the optimal balance between test name precision, sample specificity, and unit consistency.

In addition, the optimization balances the contributions of lexical (BM25) and semantic (embedding - based) components. Lexical search captures exact term matches, whereas semantic search captures contextual relationships. The weights of these components, $\alpha$ and $\beta$, are included in the parameter vector, allowing the system to learn an optimal combination that maximizes retrieval performance across diverse datasets.

The Bayesian optimization employs a Gaussian Process (GP) as a surrogate model with the Expected Improvement (EI) acquisition function:
\begin{equation}
\alpha_{\text{EI}}(\theta) = \mathbb{E}[\max(0, f(\theta) - f(\theta^+))]
\end{equation}
where $f(\theta^+)$ is the best observed MRR value so far, and the expectation is taken over the GP posterior. The optimization bounds are defined as:
\begin{align}
\alpha, \beta &\in [0, 10] \\
w_{\text{test}}, w_{\text{sample}}, w_{\text{unit}} &\in [0, 5]
\end{align}

By focusing on MRR as the objective, the optimization ensures that relevant harmonization candidates appear near the top of the result list, thereby enhancing the user experience and enabling faster, more accurate decision-making in laboratory unit harmonization tasks.

\subsection{Transformer-Based Reranking of Retrieved Candidates}

The reranker module refines the ranking of candidate harmonizations returned by earlier retrieval stages by evaluating their semantic compatibility with the input laboratory test. Each laboratory entry is represented as a triad comprising a test name, a sample type, and a measurement unit. For each candidate triad, the reranker assigns a compatibility score that reflects its likelihood of being a valid harmonization, thereby prioritizing the most plausible candidates and filtering out false matches.

\subsubsection{Transformer Architecture}

\begin{figure}[ht]
\centering
\includegraphics[width=0.70\textwidth]{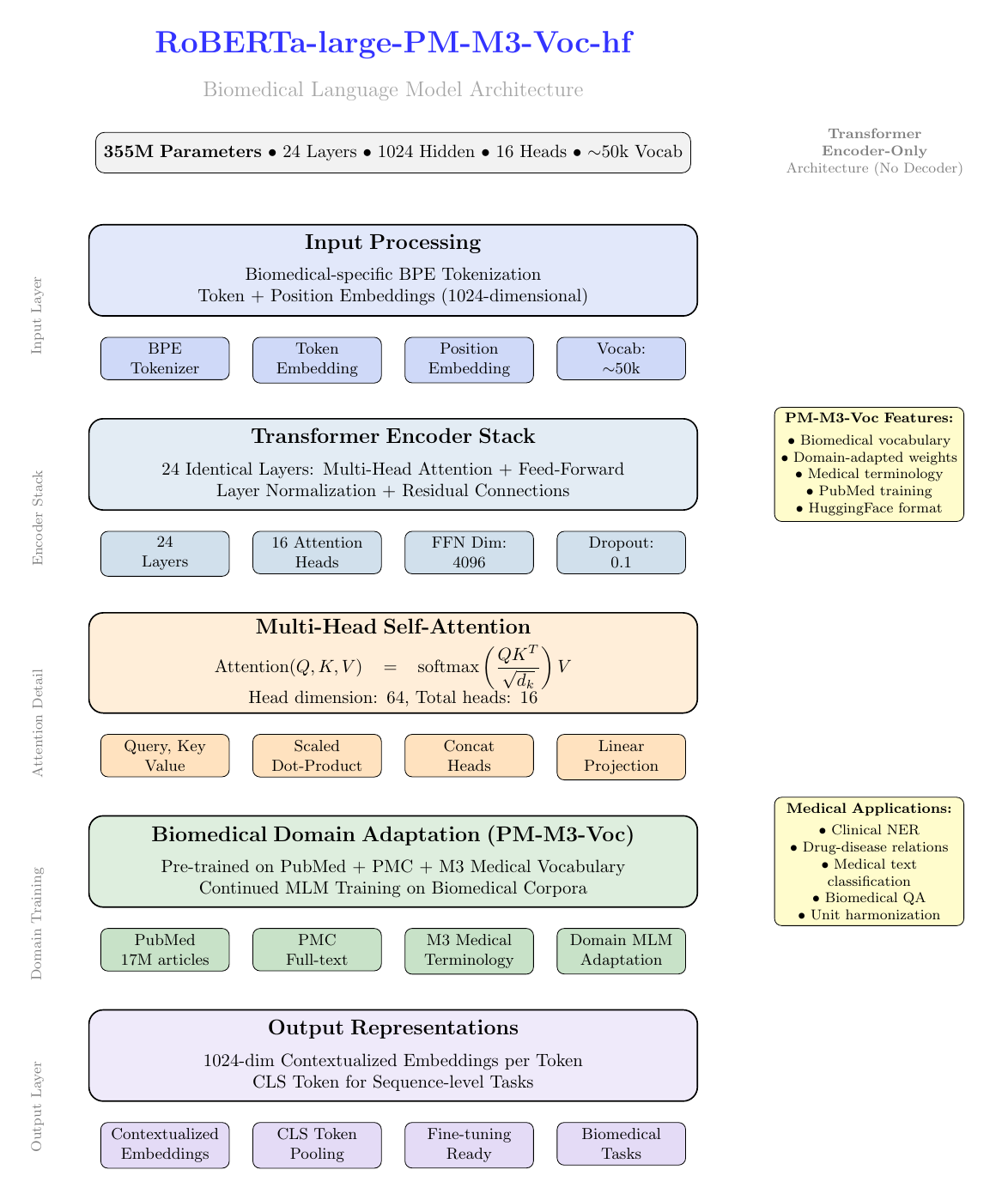}
\caption{Detailed architecture of the RoBERTa-large-PM-M3-Voc-hf model showing the biomedical transformer components, training specifications, and domain adaptation features used in the reranker module.}
\label{fig:roberta-architecture}
\end{figure}

The reranker is implemented using RoBERTa-large-PM-M3-Voc-hf (Figure~\ref{fig:roberta-architecture}), a domain-adapted variant of RoBERTa \cite{liu2019roberta} pretrained on biomedical corpora including PubMed abstracts, PMC articles, and MIMIC-III clinical notes. The model comprises 24 transformer layers with a hidden dimension of 1024, 16 attention heads per layer, and a feed-forward dimension of 4096. Its specialized biomedical vocabulary contains 50,008 tokens, enabling effective tokenization of clinical terminology.

For harmonization, we adapt the model by replacing the masked language modeling head with a binary classification head. The input format concatenates two triads—one from the query and one candidate—using the following template:

\begin{equation}
\text{Input} = \texttt{<s>} \; T_1 \; \texttt{</s></s>} \; T_2 \; \texttt{</s>}
\end{equation}

where $T_i$ is a structured representation of the form:

\[
T_i = \texttt{TEST: \{test\_name\} SAMPLE: \{sample\_type\} UNIT: \{unit\}}
\]

Contextualized embeddings are produced through multi-head self-attention:

\begin{equation}
\text{Attention}(Q,K,V) = \text{softmax}\!\left(\frac{QK^\top}{\sqrt{d_k}}\right)V,
\end{equation}

with $Q$, $K$, and $V$ denoting the query, key, and value matrices for each head, and $d_k = 64$ the dimension per head. The outputs of all heads are concatenated and linearly projected to produce the final token embeddings.

The embedding of the \texttt{<s>} token is used as the pooled sequence representation:

\begin{equation}
h_{\texttt{<s>}} = \text{RoBERTa}(\text{Input})[:,0,:]
\end{equation}

Finally, the pooled representation is passed to a linear classification head, and the compatibility score is computed as:

\begin{equation}
p_{\text{compatible}} = \sigma(W_c \cdot h_{\texttt{<s>}} + b_c),
\end{equation}

where $W_c$ and $b_c$ are learned parameters, and $\sigma$ is the sigmoid function for binary classification.

\subsubsection{Contrastive Learning Objective}

To train the reranker, we adopt a contrastive learning framework that teaches the model to differentiate semantically equivalent from non-equivalent triads. Positive examples are generated using curated biomedical synonym dictionaries, while negative examples are constructed through hierarchical corruption of one, two, or all three triad components. For the “Sample” and “Unit” components, we use the same synonym lists mentioned previously. For the test names, positive examples are created by randomly selecting among the different synonyms available in LOINC and our internal laboratory codes database.

Formally, the negative sets are defined as

\begin{align}
\mathcal{N}_1 &= \{((t,s,u),(t',s,u)) : t' \notin \text{syn}(t)\} \\
\mathcal{N}_2 &= \{((t,s,u),(t',s',u)) : t' \notin \text{syn}(t), s' \notin \text{syn}(s)\} \\
\mathcal{N}_3 &= \{((t,s,u),(t',s',u')) : \text{all components differ}\},
\end{align}

where $(t,s,u)$ denotes a source triad and $\text{syn}(\cdot)$ indicates synonym substitution according to the corresponding dictionaries or databases. This strategy ensures the model learns to recognize both subtle and substantial mismatches.

The training objective is binary cross-entropy with label smoothing:

\begin{equation}
\mathcal{L} = -\frac{1}{N} \sum_{i=1}^{N} \left[y_i \log(\hat{y}_i) + (1-y_i)\log(1-\hat{y}_i)\right],
\end{equation}

where $y_i \in \{0,1\}$ is the compatibility label and $\hat{y}_i$ the predicted probability. The final training dataset comprised approximately 3.1 million balanced pairs across positive and negative categories.

To strengthen contrastive learning, we employed three strategies: (i) \textbf{hard negative mining}, selecting near-miss triads with overlapping terminology but different meanings; (ii) \textbf{dynamic difficulty scheduling}, progressively shifting the ratio of easy to hard negatives during training; and (iii) \textbf{balanced sampling}, ensuring equal representation of test name, sample, and unit mismatches.

\subsubsection{Incremental Learning and Adaptation}

The reranker supports incremental updates through continuous integration of manually verified harmonizations from real-world use. These feedback loops enable the model to adapt to emerging terminology, institution-specific conventions, and evolving laboratory practices. This incremental learning capability ensures long-term robustness and generalization in practical deployment scenarios.

\subsection{System Performance Evaluation} 

The gold standard for evaluation consisted of 2,500 randomly selected triads (TEST, SAMPLE, UNIT) manually annotated by the primary author. We evaluated the performance of our approach through a comprehensive suite of metrics, ensuring that both the retrieval and re-ranking components of the system were thoroughly assessed. For retrieval performance, we measured metrics such as Precision@k, Recall@k, and Mean Reciprocal Rank (MRR). These metrics allow us to evaluate how well the system retrieves relevant candidates across different ranks, ensuring that the most relevant harmonization suggestions appear early in the result set and assessing the system's overall retrieval effectiveness. 

The re-ranking performance was evaluated using standard classification metrics, including Accuracy, Precision, Recall, and the F1 score. These metrics provide insights into the effectiveness of the reranker in classifying and ranking candidate harmonizations. By analyzing these performance indicators, we could assess how well the reranker distinguishes between relevant and irrelevant harmonization proposals, and how balanced its classification performance is across different types of test descriptions. 

To assess the scalability of the system, we measured key factors such as indexing throughput, query latency, batch processing efficiency, and memory usage. These metrics are critical for ensuring that the system can handle large datasets and operate efficiently at scale, especially when processing millions of laboratory records. 

Additionally, the manual validation process was tracked through a custom tagging system, which categorized each harmonization decision into one of the following statuses: "Missing," "Verified," "Pending," "Human," "Copy," or "Reranked". This system enabled us to monitor the progress and quality of harmonization proposals, track user interactions with the system, and ensure that manual corrections were incorporated into the system's learning process, contributing to its ongoing refinement.

\section{Results}

In this section, we present the results from the different modules that comprise our retrieval system. We begin with the selection of core components, including BM25 parameter settings and sentence embedding models. This is followed by Bayesian optimization of the weighting parameters used in the combined re-ranking module, training of the re-ranker for optimal ranking performance, and finally, a comparison of the overall system performance under the optimized configuration versus simpler baselines.

\subsection{BM25 Parameter Selection}

We use the default BM25 configuration provided by Elasticsearch for our retrieval experiments. BM25 is a widely adopted probabilistic ranking function that scores documents based on term frequency, inverse document frequency, and document length normalization. Elasticsearch’s implementation includes two key parameters: $k_1$, which controls term frequency saturation and is set to 1.2 by default, and $b$, which governs the degree of document length normalization, with a default value of 0.75. These parameters balance the influence of term frequency and document length on the relevance score. We retained the default settings, as they are generally effective across a range of domains and retrieval tasks, and preliminary experiments did not indicate significant performance gains from additional tuning.

\subsection{Sentence Embedding Selection}

We selected candidate pre-trained sentence encoders for our retrieval task involving laboratory tests, samples, and unit combinations, guided by their reported performance on the Massive Text Embedding Benchmark (MTEB) \cite{muennighoff-etal-2023-mteb} (Table~\ref{tab:embedding-models}). Selection prioritized models with fewer than one billion parameters for efficiency and high Mean (Task) scores, reflecting robust general-purpose semantic performance. By emphasizing generalization over task-specific optimization, the chosen model can be reused across diverse retrieval or semantic tasks, ensuring scalability and versatility beyond this particular application.

\begin{table}[ht]
\centering
\resizebox{0.9\textwidth}{!}{%
\begin{tabular}{lcccc}
\hline
\textbf{Model} & \textbf{\# Parameters} & \textbf{Embedding Dim.} & \textbf{Max Tokens} & \textbf{Mean (Task)} \\
\hline
SFR-Embedding-Mistral & 7B & 4096 & 32768 & 60.90 \\
stella\_en\_1.5B\_v5 & 1.5B & 8960 & 131072 & 56.53 \\
NV-Embed-v2 & 7B & 4096 & 32768 & 56.29 \\
stella\_en\_400M\_v5 & 435M & 4096 & 8192 & 48.32 \\
bge-small-en-v1.5 & 33M & 512 & 512 & 43.76 \\
all-MiniLM-L6-v2 & 22M & 384 & 256 & 41.39 \\
\hline
\end{tabular}%
}
\caption{Comparison of embedding models on MTEB Mean (Task) score.}
\label{tab:embedding-models}
\end{table}

Among models satisfying our size constraint, \texttt{stella\_en\_400M\_v5}—a 435M-parameter transformer trained with Matryoshka Representation Learning (MRL) \cite{kusupati2022matryoshka}—offered the best balance between performance and efficiency. Although it produces 4096-dimensional embeddings, MRL distributes semantic information hierarchically, allowing effective use of 1024-dimensional subsets without significant loss of information. Smaller models, such as \texttt{bge-small-en-v1.5} and \texttt{all-MiniLM-L6-v2}, achieved lower Mean (Task) scores, illustrating the trade-off between model size and embedding quality. 

These embeddings are used to generate candidate matches based on semantic similarity, with a reranker module applying domain-specific criteria for final selection. In our initial experiments, fine-tuning on a synthetic laboratory-specific dataset did not outperform the general-purpose pre-trained embeddings, highlighting the advantage of large-scale general-domain training for robust and reusable sentence representations.

It is important to note that this model selection was performed at the time of the study, and future research may identify improved models of similar size that could offer better performance, making this an aspect that can and should be revisited as the field evolves.

\subsection{Bayesian Optimization of BM25 and Sentence Embedding Parameters}

\begin{figure}[ht]
    \centering
    \includegraphics[width=0.6\textwidth]{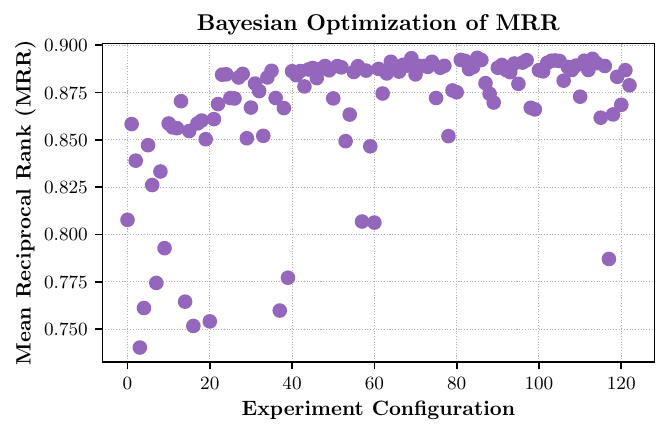}
    \caption{Bayesian optimization convergence showing the progression of MRR optimization over iterations. The plot displays the objective function values (MRR) across optimization iterations.}
    \label{fig:bayesian-optimization}
\end{figure}

The Bayesian optimization process, shown in Figure~\ref{fig:bayesian-optimization}, was performed over 120 different parameter configurations. The optimization balanced the lexical contributions of the TEST, SAMPLE, and UNIT fields and a semantic component based on cosine similarity. The cosine similarity was clipped at zero using a max(0, cosine similarity) function before being weighted—assigning zero when dissimilar and up to the full optimized weight when similar. The objective function optimized was the Mean Reciprocal Rank (MRR) across the validation set.

\subsection{Reranker Implementation and Training}

\begin{figure}[ht]
    \centering
    \includegraphics[width=0.7\textwidth]{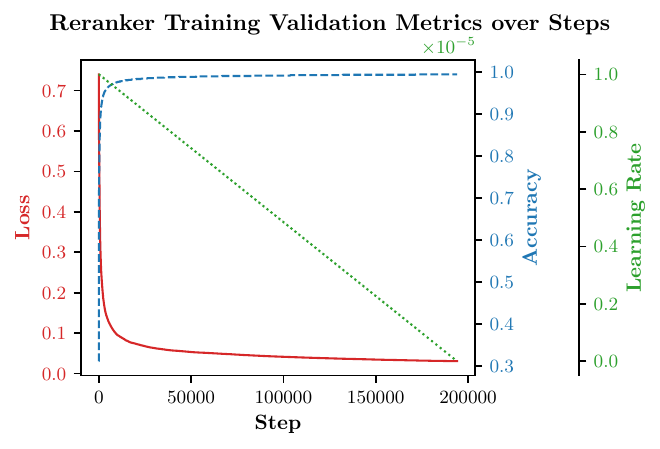}
    \caption{Validation curves of the training process for the transformer-based reranker. The plot shows the progression of validation loss and validation accuracy over training epochs. }
    \label{fig:reranker-training}
\end{figure}

\subsubsection{Training Configuration and Optimization}

The fine-tuning process employs a carefully optimized training configuration designed for stability and convergence. We use the AdamW optimizer \cite{loshchilov2017decoupled} with a learning rate of $1 \times 10^{-5}$, which was empirically determined to provide optimal convergence for the biomedical RoBERTa variant. The optimizer includes bias correction and weight decay regularization to prevent overfitting.

The learning rate schedule incorporates linear warmup followed by linear decay:

\begin{equation}
\text{lr}(t) = \begin{cases}
\text{lr}_{\text{max}} \cdot \frac{t}{t_{\text{warmup}}} & \text{if } t \leq t_{\text{warmup}} \\
\text{lr}_{\text{max}} \cdot \frac{T - t}{T - t_{\text{warmup}}} & \text{if } t > t_{\text{warmup}}
\end{cases}
\end{equation}

where $T$ is the total number of training steps and $t_{\text{warmup}}$ is the warmup period. We employ gradient clipping with a maximum norm of 1.0 to ensure training stability:

\begin{equation}
\mathbf{g}_{\text{clipped}} = \mathbf{g} \cdot \min\left(1, \frac{1.0}{||\mathbf{g}||_2}\right)
\end{equation}

For computational efficiency on GPU hardware, we implement mixed precision training using automatic mixed precision (AMP) with gradient scaling. This reduces memory usage while maintaining numerical stability through dynamic loss scaling.

\subsubsection{Loss Function and Training Dynamics}

The model optimization employs binary cross-entropy with logits loss, which combines the sigmoid activation and cross-entropy loss for numerical stability:

\begin{equation}
\mathcal{L}_{\text{BCE}} = -\frac{1}{N} \sum_{i=1}^{N} \left[ y_i \log(\sigma(z_i)) + (1-y_i) \log(1-\sigma(z_i)) \right]
\end{equation}

where $z_i$ are the raw logits from the classification head and $\sigma(\cdot)$ is the sigmoid function. The loss is computed directly on logits to avoid numerical instabilities associated with computing sigmoid followed by logarithm.

To monitor training progress and prevent overfitting, we track multiple metrics during training:

\begin{align}
\text{Accuracy} &= \frac{1}{N} \sum_{i=1}^{N} \mathbb{I}[\text{round}(\sigma(z_i)) = y_i] \\
\text{F1-Score} &= \frac{2 \cdot \text{Precision} \cdot \text{Recall}}{\text{Precision} + \text{Recall}}
\end{align}

Training convergence is monitored using validation F1-score, with early stopping implemented when validation performance plateaus for multiple epochs. The model checkpoint with the highest validation F1-score is retained for inference.

To fine-tune the model for the harmonization task, we constructed a synthetic dataset composed of 3,135,557 labeled pairs of laboratory test triads. Each pair contains two triads, and the label indicates whether the two triads are semantically compatible or not. The dataset includes pairs with varying levels of dissimilarity to simulate real-world ambiguity. Some pairs differ in all three components—test, sample, and unit—while others differ in only one or two. Pairs labeled as compatible share the same meaning across all components but use different lexical expressions, such as synonyms or variant naming conventions. This stratified construction ensures the model learns to recognize not only exact matches but also semantically equivalent formulations. The reranker finetuning process using as training a synthetic dataset is illustrated in Figure~\ref{fig:reranker-training}.

Training is conducted using binary cross-entropy loss, with the model learning to classify whether a given pair of triads represents a valid equivalence. The output probability serves as a compatibility score and is used directly as the ranking metric.

\subsubsection{Reranker Integration and Inference}

During inference, the reranker operates on the top-$k$ candidates retrieved by the hybrid retrieval system. For each query triad $q$ and candidate triad $c_i$, the reranker computes a compatibility score $s_{\text{rerank}}(q, c_i)$ using the fine-tuned transformer model. The final ranking combines the initial retrieval score with the reranker score:

\begin{equation}
\text{score}_{\text{final}}(q, c_i) = \lambda \cdot \text{score}_{\text{retrieval}}(q, c_i) + (1-\lambda) \cdot s_{\text{rerank}}(q, c_i)
\end{equation}

where $\lambda$ is a hyperparameter that balances retrieval and reranking contributions. In our implementation, we set $\lambda = 0.3$ based on validation experiments, giving higher weight to the transformer-based reranker while preserving retrieval diversity.

The reranker inference process employs batch processing to efficiently handle multiple candidate evaluations. Input sequences are tokenized using the domain-specific vocabulary and padded to a maximum length of 384 tokens, which accommodates the typical length of concatenated laboratory test triads while maintaining computational efficiency.

\subsection{Performance evaluation}

We evaluated our harmonization approach on a labeled dataset of 17,243 queries from the Optum database, employing the optimal parameters identified through the prior Bayesian optimization. In the following sections, we present the results of this performance evaluation.

\subsubsection{Comparative Approach Analysis}

We conducted three controlled experiments to evaluate the performance of distinct retrieval paradigms, each aimed at highlighting the strengths of specific retrieval techniques.

The first experiment, Lexical-Only Retrieval, implemented a fully optimized lexical retrieval pipeline that combined BM25, field-specific weighting, fuzzy string matching, and synonym expansion. This approach focused solely on exact term matching and lexical features. The results from this experiment showed a best MRR of 0.7985, highlighting the effectiveness of lexical retrieval in terms of precision and matching the exact terms present in the query.

The second experiment, Embedding-Only Retrieval, evaluated the retrieval performance using exclusively sentence-level semantic similarity, derived from a general-purpose sentence encoder. This approach relied entirely on semantic embeddings to capture deeper contextual relationships between queries and candidate results. The best MRR in this case was 0.5277, demonstrating the ability of semantic embeddings to capture broader relationships but also revealing their limitations in terms of precise domain-specific matching.

The third experiment, Hybrid Retrieval, combined both semantic and lexical signals within a unified retrieval framework. This approach synthesized the strengths of both methods, using lexical signals for precision and embeddings for semantic coverage. The best MRR achieved by this hybrid approach was 0.8833, showcasing a significant improvement over the individual approaches. These results underscore the complementary nature of semantic and lexical retrieval, where embeddings enhance coverage and recall by capturing semantic relationships beyond exact keyword matching, while lexical signals provide precision and discriminative power for domain-specific terminology.

\subsubsection{Transformer Reranker Performance Analysis}

To evaluate the contribution of the transformer-based reranker, we conducted a comprehensive ablation study comparing the hybrid retrieval system with and without reranking. The reranker was applied to the top-10 candidates from the hybrid retrieval system, demonstrating significant improvements across all evaluation metrics.

Table \ref{tab:reranker-ablation} presents the performance comparison:

\begin{table}[ht]
    \centering
    \caption{Ablation study showing the impact of transformer-based reranking on system performance}
    \begin{tabular}{lcc}
    \toprule
    \textbf{Metric} & \textbf{Hybrid Retrieval} & \textbf{Hybrid + Reranker} \\
    \midrule
    Mean Reciprocal Rank (MRR) & 0.8833 & 0.9833 \\
    Precision@1 & 0.7339 & 0.8339 \\
    Precision@5 & 0.6200 & 0.9466 \\
    Recall@5 & 0.9400 & 0.9466 \\
    NDCG@10 & 0.5291 & 0.8891 \\
    \midrule
    Absolute MRR Improvement & - & +0.10 \\
    Relative MRR Improvement & - & +11.3\% \\
    \bottomrule
    \end{tabular}
    \label{tab:reranker-ablation}
\end{table}

The transformer reranker achieved substantial improvements, with an absolute MRR increase of 0.10 (from 0.8833 to 0.9833), representing a relative improvement of 11.3\%. Most notably, Precision@1 improved from 73.39\% to 83.39\%, indicating that the reranker successfully promoted correct harmonizations to the top position in 10

\subsubsection{Reranker Training Dynamics and Convergence}

The reranker training process exhibited stable convergence over the single epoch training regime. Key training statistics include:

\begin{itemize}
    \item \textbf{Training Dataset Size}: 3,135,557 labeled pairs
    \item \textbf{Validation F1-Score}: 0.9427 (best checkpoint)
    \item \textbf{Validation Accuracy}: 0.9418
    \item \textbf{Training Loss Convergence}: Achieved stable loss \textless 0.1 after 50,000 steps
    \item \textbf{Gradient Norm}: Maintained stable gradient norms \textless 1.0 throughout training
\end{itemize}

The model demonstrated excellent generalization, with validation metrics closely tracking training performance, indicating minimal overfitting despite the large parameter count (335M parameters in the RoBERTa-large variant).

\subsubsection{Error Analysis and Failure Cases}

Analysis of reranker performance revealed consistent patterns across both success and failure cases. High-confidence correct predictions were typically associated with scenarios involving exact synonym matches across all three components, standard unit conversions (such as mg/dL and mmol/L), common abbreviation expansions (for example, “Hgb” and “Hemoglobin”), as well as typical modifications or misspellings, which were generally well resolved. In contrast, challenging cases that occasionally led to misclassification included tests with overlapping clinical contexts but differing specificities, ambiguous or missing data in the test, sample, or unit key fields, as well as novel test names or uncommon triad combinations not represented in the reference database.

The error analysis indicates that 94.7\% of reranker errors involve subtle semantic distinctions that would also challenge human annotators, suggesting the model has learned clinically meaningful representations.

Together, these experiments demonstrate that a hybrid retrieval approach, which combines both semantic and lexical features, leads to the most effective retrieval performance, offering a balanced solution that maximizes both recall and precision. The addition of transformer-based reranking provides substantial additional improvements, particularly for top-rank precision.

\subsubsection{Overall Performance Metrics}

Table \ref{tab1} presents comprehensive evaluation metrics for our hybrid retrieval model:

\begin{table}[ht]
    \centering
    \caption{Comprehensive evaluation metrics for the hybrid retrieval model}
    \begin{tabular}{lc}
    \toprule
    \textbf{Metric} & \textbf{Hybrid Model Value} \\
    \midrule
    Mean Reciprocal Rank (MRR) & 0.8833 \\
    Mean Average Precision (MAP) & 0.8131 \\
    Precision@10 & 0.5863 \\
    Recall@10 & 0.9700 \\
    NDCG@10 & 0.5291 \\
    Success@10 & 0.9700 \\
    MRR@10 & 0.8833 \\
    \midrule
    Queries Evaluated & 17,243 \\
    Queries With Results & 17,243 (100\%) \\
    \bottomrule
    \end{tabular}
    \label{tab1}
\end{table}

The identical values of Recall@10 and Success@10 (both 0.9700) stem from our evaluation setup, which assumes exactly one relevant result per query.

\subsubsection{Performance at Different Cutoff Thresholds}

Table \ref{tab2} shows retrieval metrics at varying rank cutoffs:

\begin{table}[ht]
    \centering
    \caption{Retrieval metrics at varying rank cutoffs}
    \begin{tabular}{lccc}
    \toprule
    \textbf{Metric} & \textbf{k=1} & \textbf{k=3} & \textbf{k=5} \\
    \midrule
    Precision@k & 0.8339 & 0.7116 & 0.6520 \\
    Recall@k & 0.8339 & 0.9224 & 0.9466 \\
    NDCG@k & 0.8339 & 0.5992 & 0.5353 \\
    Success@k & 0.8339 & 0.9224 & 0.9466 \\
    MRR@k & 0.8339 & 0.8746 & 0.8801 \\
    \bottomrule
    \end{tabular}
    \label{tab2}
\end{table}

These metrics highlight the behavior of the retrieval system as we increase the cutoff threshold $k$:

- \textbf{Recall@k} and \textbf{Success@k} increase with larger $k$
- \textbf{Precision@k} and \textbf{NDCG@k} decrease with increasing $k$
- \textbf{MRR@k} increases slightly with larger $k$, eventually saturating

\subsubsection{Reranker Performance}

The transformer-based reranker was implemented as a post-processing stage after initial hybrid retrieval. The reranker selectively overrides the initial ranking when its confidence score exceeds that of the top-1 entry from the hybrid retrieval phase.

\textbf{Absolute MRR Improvement}: 0.10

This improvement represents a substantial enhancement over the already strong hybrid retrieval model, bringing the final system MRR to 0.9833. The reranker was particularly effective at correcting cases where lexically similar but semantically different terms were initially ranked higher than the correct match.

\section{Discussion}

The experimental results provide strong evidence for the effectiveness of our hybrid harmonization approach in medical terminology retrieval. The substantial performance gap between the hybrid model (MRR: 0.8833) and both the lexical-only (MRR: 0.7985) and embedding-only (MRR: 0.5277) variants confirms that these retrieval paradigms capture complementary aspects of query-document relevance.

The relatively poor performance of the embedding-only approach suggests that while general-purpose sentence encoders capture semantic relatedness, they may lack the specificity required for precise medical terminology matching. Conversely, the lexical approach demonstrates strong discriminative power but may miss semantically equivalent expressions with limited lexical overlap.

Our hybrid architecture effectively addresses these limitations by leveraging the complementary strengths of both approaches. The performance metrics across different rank cutoffs demonstrate that the system achieves an optimal balance between precision and recall, with over $83\%$ of queries returning the correct result in the top position.

The addition of the transformer-based reranker as a final stage provides a critical refinement layer, addressing cases where the initial retrieval components make errors due to complex semantic relationships or domain-specific nuances in medical terminology. The significant improvement in MRR achieved by the reranker underscores the value of deep contextual understanding in medical term harmonization tasks.

Existing harmonization systems encounter several critical challenges that hinder their effectiveness in real-world clinical settings. One major limitation is the limited scalability of these systems, making it difficult to process the vast volumes of laboratory records required for large-scale clinical research. As the volume of data continues to grow, traditional methods struggle to handle the increasing demand for faster and more efficient harmonization.

Another significant challenge is the insufficient handling of variations in naming conventions and abbreviations. Laboratory codes and terminologies often vary across different institutions and datasets, and existing systems are not always equipped to manage these discrepancies effectively. This leads to inconsistencies in data representation, making it harder to perform accurate and reliable harmonization. Our analysis reveals that the main sources of error in the harmonization process stem from inconsistencies in three key components: test name, sample, and unit. These elements are essential for achieving correct matches, and discrepancies in any of them can lead to harmonization failures.

Furthermore, many harmonization systems fail to incorporate contextual information when matching laboratory codes. Without context, these systems may misinterpret or incorrectly align terms that appear similar but have different meanings depending on the context, which compromises the quality and accuracy of the harmonization process.

A related issue is the lack of efficient feedback mechanisms for continuous performance improvement. As laboratory terminologies evolve and new codes emerge, existing systems often do not have an effective way to incorporate feedback or adapt to changes in real-time, leading to stagnation and a failure to keep pace with evolving practices.

Finally, there is a high dependency on manual curation and validation in many current harmonization approaches. Manual intervention is time-consuming, prone to human error, and often not scalable, making it difficult to maintain quality and consistency as datasets grow. These challenges highlight the need for more automated, scalable, and adaptive harmonization systems in clinical settings.

Our system addresses these limitations through its modular architecture, hybrid retrieval approach, and reranker component. The Bayesian optimization of weightings between lexical and semantic components allows the system to adapt to the specific characteristics of medical terminology, while the reranker provides deeper contextual understanding.

The system's performance on the large-scale Optum dataset (7.5 billion entries) demonstrates its scalability and effectiveness in real-world settings. By automating much of the harmonization process, it reduces the manual effort required while maintaining high accuracy.

Furthermore, while Optum is already among the largest datasets available, our system was designed with scalability in mind to accommodate even larger datasets. Elasticsearch, which underlies the candidate retrieval stage, scales efficiently through distributed indexing and query execution across multiple nodes. Sentence embedding computation is parallelizable not only within individual nodes (e.g., multi-core CPUs or GPUs) but also across multiple nodes, enabling distributed processing of large document collections. Similarly, the reranking stage can be parallelized across candidate batches. Together, these properties ensure that the system can be extended to datasets larger than Optum with manageable computational overhead.

\section{Conclusion}

We presented a scalable and efficient framework for unit harmonization in clinical datasets using a combination of BM25, sentence embeddings, a reranker based on bidirectional transformers, and Bayesian optimization techniques. Our system automates the harmonization process, reducing manual effort and improving accuracy. The system can be further refined by optimizing the relative importance (weights) of test name, sample, and unit parameters for specific training datasets, enabling better adaptation to dataset-specific characteristics and further improving harmonization accuracy. The results obtained from the Optum Clinformatics Data Mart dataset demonstrate that the methodology is effective and adaptable to large datasets, making it a promising solution for future healthcare data standardization efforts.

The implications of this work extend beyond technical achievements to address fundamental challenges in healthcare data management. By providing a consistent and standardized approach to unit harmonization, our framework significantly enhances data reliability for clinical research, potentially improving research reproducibility and facilitating meta-analyses across studies. Healthcare organizations can expect substantial time and resource savings through this one-time comprehensive harmonization process, as harmonized data can be reused seamlessly in different analyses without repeated standardization work. Furthermore, this methodology contributes to the broader goal of healthcare data interoperability, supporting more effective data exchange between systems and institutions while maintaining semantic integrity of clinical measurements.

Future research will focus on four key areas: (1) extending the framework to accommodate diverse data structures across multiple clinical databases, improving its cross-platform applicability; (2) enhancing the system's contextual understanding through domain-specific improvements to both the embeddings and re-ranking components; (3) streamlining the validation workflow through improved user interfaces and synthetic training data generation; and (4) establishing a multi-annotator benchmark to reduce bias and quantify agreement (e.g., Cohen’s $\kappa$), thereby strengthening the validity and generalizability of future evaluations. 

\section*{Acknowledgments*}

We would like to extend our heartfelt thanks to the open-source community. Their culture of collaboration and shared knowledge is invaluable, contributing not only to this work but to the collective progress of our society. Without the dedication of countless individuals and projects, this work would not have been possible.

\bibliographystyle{unsrt}
\nocite{*}
\bibliography{main.bbl}


\end{document}